\documentclass[11pt]{article}
\usepackage{coling2020}
\usepackage{times}
\usepackage{url}
\usepackage{latexsym}
\usepackage{xcolor,booktabs,graphicx,amsmath,amssymb,tablefootnote}

\colingfinalcopy

\newcommand{\citet}{\newcite}

\newcommand{\MC}[3]{\multicolumn{#1}{#2}{#3}}

\newcommand{\vatex}{\textsc{VaTeX}}
\newcommand{\sd}[1]{\small{$\pm$ #1}}
\newcommand{\B}[1]{\textbf{#1}}
\newcommand{\I}[1]{\textit{#1}}
\newcommand{\T}[1]{\texttt{#1}}
\newcommand{\da}[1]{\textcolor{purple}{\small $\Downarrow {#1}$}}

\newcommand{\ie}[1]{\textit{i.e.}~#1}
\newcommand{\eg}[1]{\textit{e.g.}~#1}

\definecolor{tableh}{RGB}{128, 128, 128}
\definecolor{grayrow}{RGB}{100, 100, 100}
\newcommand{\tableh}[1]{\textcolor{tableh}{#1}}

\newcommand{\bss}[1]{``\textit{\textcolor{black}{#1}}''}
\newcommand{\BSSCocoEn}{\bss{A pizza is sitting on top of a table}}
\newcommand{\BSSCocoJa}{\bss{A person who is skiing stands}}

\newcommand{\BSSFlickr}{\bss{A man in a blue shirt standing in front of a building}}
\newcommand{\BSSMSVD}{\bss{A man is playing}}
\newcommand{\BSSVatexEn}{\bss{A man is working out in front of a group of people}}
\newcommand{\BSSVatexZh}{\bss{In a room, a man in black is dancing}}
\newcommand{\BSSDDialog}{\bss{Where do you want to go?}}

\title{Curious Case of Language Generation Evaluation Metrics: \\A Cautionary Tale}

\author{Ozan Caglayan$^{1}${\normalfont,} Pranava Madhyastha$^1$ \and Lucia Specia$^{1,2,3}$\\[.3em]
Imperial College London$^1$,\, University of Sheffield$^2$,\, ADAPT - Dublin City University$^3$\\
\texttt{\small o.caglayan@ic.ac.uk, pranava@ic.ac.uk, l.specia@ic.ac.uk}\\
}

\date{}

\begin{document}
\maketitle

\begin{abstract}
Automatic evaluation of language generation systems is a well-studied problem in Natural Language Processing. While novel metrics are proposed every year, a few popular metrics remain as the \I{de facto} metrics to evaluate tasks such as image captioning and machine translation, despite their known limitations. This is partly due to ease of use, and partly because researchers expect to see them and know how to interpret them. In this paper, we urge the community for more careful consideration of how they automatically evaluate their models by demonstrating important failure cases on multiple datasets, language pairs and tasks. Our experiments show that metrics (i) usually prefer system outputs to human-authored texts, (ii) can be insensitive to correct translations of rare words, (iii) can yield surprisingly high scores when given a single sentence as system output for the entire test set.
\end{abstract}

\section{Introduction}
\label{sec:intro}
Human assessment is the best practice at hand for evaluating language generation tasks such as machine translation (MT), dialogue systems, visual captioning and abstractive summarisation. In practice, however, we rely on automatic metrics which compare system outputs to human-authored references. Initially proposed for MT evaluation, metrics such as \textsc{Bleu}~\cite{papineni-etal-2002-bleu} and \textsc{Meteor}~\cite{denkowski-lavie-2014-meteor} are increasingly used for other tasks, along with more task-oriented metrics such as \textsc{Rouge}~\cite{lin-2004-rouge} for summarisation and \textsc{Cider}~\cite{vedantam2015cider} for visual captioning.

\citet{reiter-belz-2009-investigation} remark that
it is \I{not sufficient} to conclude on the usefulness of a natural language generation system's output by solely relying on metrics that quantify the similarity of the output to human-authored texts.
Previous criticisms concerning automatic metrics corroborate this perspective to some degree. To cite a few, in the context of MT, 
\citet{callison-burch-etal-2006-evaluation} state that human judgments may not correlate with \textsc{Bleu} and an increase in \textsc{Bleu} does not always indicate an improvement in quality. Further, \citet{mathur-etal-2020-tangled} challenge the stability of the common practices measuring correlation between metrics and human judgments, the standard approach in the MT community, and show that they may be severely impacted by outlier systems and the sample size.
In the context of image captioning, \citet{Wang_2019_CVPR} claim that the consensus-based evaluation protocol of \textsc{Cider} actually penalises output diversity. Similar problems have also been discussed in the area of automatic summarisation with respect to~\textsc{Rouge}~\cite{schluter-2017-limits}.
Nevertheless, automatic metrics like these are a necessity and remain popular, especially given the increasing number of open evaluation challenges.

In this paper, we 
further probe \textsc{Bleu, Meteor, Cider-D} and \textsc{Rouge}$_\text{L}$ metrics ($\S$~\ref{sec:metrics}) that are commonly used to quantify progress in language generation tasks. We also probe a recently proposed contextualised embeddings-based metric called \textsc{BertScore}~\cite{bertscore}. We first conduct \textit{leave-one-out} average scoring with multiple references and show that, counter-intuitively, metrics tend to reward system outputs more than human-authored references ($\S$~\ref{sec:humaneval}).
A system output perturbation experiment further highlights how metrics penalise errors in extremely frequent $n$-grams ($\S$~\ref{sec:freqpert}) while
they are quite insensitive to errors in rare words ($\S$~\ref{sec:infreq}).
The latter makes it hard to understand whether a model is better than another in its ability to handle rare linguistic phenomena correctly.
Finally, we design ($\S$~\ref{sec:bss}) an adversary that seeks to demonstrate metrics' preference for frequent $n$-grams by using a single training set sentence as system output for the entire test set. We observe strikingly high scores, such as the sentence \BSSMSVD\, obtaining a \textsc{Bleu} score of $30.6$, compared to $47.9$ of a strong model, on a captioning task.
We hope that our observations in this paper will lead the community towards formulation of better metrics and evaluation protocols in the future ($\S$~\ref{sec:discussions}).

\section{Automatic Metrics}
\label{sec:metrics}
In this section, we briefly describe the metrics we use in our experiments. To compute \textsc{Bleu}, \textsc{Meteor}, \textsc{CIDEr} and \textsc{Rouge-L}, we provide pre-tokenised hypotheses and references to \T{coco-caption} utility\footnote{\url{https://github.com/tylin/coco-caption}}. For \textsc{BertScore}, we use its official release\footnote{Version 0.3.4 \T{hug\_trans 2.11.0}, with default settings except \T{\small rescale-with-baseline} is enabled.}.
\begin{itemize}
  \item Initially proposed for MT evaluation, \B{\textsc{Bleu}} is a prevalent metric based on $n$-gram matches between the candidate and reference sentences. The final score is the geometric mean of $n$-gram \textit{precisions} with a brevity penalty to penalise outputs shorter than references.
  \item In contrast, \B{\textsc{Meteor}} is a \textit{recall}-based metric which uses explicit word alignments between candidate and references, allowing for exact matching, synonymy, and stemming. A fragmentation penalty rewards longer and fewer chunks of contiguously aligned tokens. The final score is computed between the \emph{best} scoring reference and the candidate sentence.
  \item Another \textit{recall}-based metric is \B{\textsc{Rouge-L}} which measures longest common sub-sequences between the candidate and references, \ie a set of shared words with similar order even if not contiguous.
  \item \B{\textsc{CIDEr}} is a consensus-based captioning metric which uses term frequency inverse document frequency (TF-IDF) for $n$-gram weighting over all references. The final score is the average cosine similarity between the candidate sentence and references. We use the popular variant \textsc{CIDEr-D}, which integrates a length-based Gaussian penalty and clipping.
  \item Finally, \B{\textsc{BertScore}} computes a token level similarity for each candidate token against each token in the reference sentence, using contextual BERT embeddings~\cite{bertscore}. We report the F-score, \ie the harmonic mean of the precision and the recall of maximal per-token cosine similarities between the candidate and the reference(s).
\end{itemize}

\section{Experiments}
\label{sec:experiments}
This section describes our probing experiments and results\footnote{\textsc{Bleu, Meteor, Cider, Rouge, BertScore} may sometimes be abbreviated as \textsc{Bl, Mt, Cr, Rg, Bs}, for readability.} for different tasks and language pairs.

\subsection{Machine vs.\ human-authored texts}
\label{sec:humaneval}
We first explore the extent to which metrics reward human-authored texts. We hypothesise that metrics should prefer human-authored texts to machine-produced texts, as the former is considered to be the `ground truth' reference. For a given \B{multi-reference} metric $\textsc{M}$,
\textsc{Ref-Vs-Ref} \I{leave-one-out}\footnote{Our approach differs from \citet{vinyals2015show} in that we go beyond \textsc{Bleu1}.} average $\textsc{L}_{\textsc{M}}$ is computed across $C$ human-authored reference corpora $\{R_1,\dots,R_C\}$. At each iteration $i$, the reference $R_i$ is taken as the candidate and is evaluated against the held-out $C\mathrm{-}1$ references:
\vspace*{-2mm}
\begin{equation}
    \textsc{L}_{\textsc{M}} = \dfrac{1}{C}\sum_{i=1}^C{\textsc{M}(\textsc{Sys}\mathrm{=}R_i, \textsc{Refs}\mathrm{=}\{R_{j}:j\ne i\})}\label{eq:loo}
    \vspace*{-2mm}
\end{equation}
Same principle holds when evaluating a trained system, except that \textsc{Sys} would now represent the system outputs. The metric computations inside the sum would still be using $C\mathrm{-}1$ references at each iteration.

\begin{table}[t]
  \parbox[t]{.54\columnwidth}{
  \centering
  \resizebox{.54\columnwidth}{!}{%
  \begin{tabular}{@{}llllll@{}}
  \\[.55cm]
    \toprule
    &
    \MC{1}{c}{\textsc{Bl}}  &
    \MC{1}{c}{\textsc{Mt}} & 
    \MC{1}{c}{\textsc{Cr}} &
    \MC{1}{c}{\textsc{Rg}} &
    \MC{1}{c}{\textsc{Bs}} \\
    \midrule
    & \MC{5}{l}{\tableh{NIST-2008 Chinese$\rightarrow$English ($C=4$)}}\\
    \textsc{RvR} & 38.2 \sd{2.2} & \B{33.1} \sd{.9} & \B{2.20} \sd{.10} & 59.5 \sd{1.8} & \B{0.69} \sd{.020} \\
    \textsc{SvR} & \B{41.8} \sd{.7} & 32.6 \sd{.3} &     2.13 \sd{.07} & 59.5 \sd{.5} & 0.68 \sd{.003} \\
    \cmidrule(l){2-6}
    & \MC{5}{l}{\tableh{COCO Image$\rightarrow$English ($C=5$)}} \\
    \textsc{RvR} &  19.5     \sd{.3} & 24.2     \sd{.1} & 0.88     \sd{.01} & 46.7     \sd{.3} & 0.55 \sd{.003} \\
    \textsc{SvR} &  \B{30.3} \sd{.8} & \B{25.1} \sd{.3} & \B{1.05} \sd{.02} & \B{53.0} \sd{.5} & \B{0.61} \sd{.004} \\
    \cmidrule(l){2-6}
    & \MC{5}{l}{\tableh{\vatex\ Video$\rightarrow$English ($C=10$)}}\\
    \textsc{RvR} & 24.1 \sd{.2}    &\B{24.9} \sd{.1} & \B{0.63} \sd{.01}& 48.8 \sd{.1}    & 0.57 \sd{.002} \\
    \textsc{SvR} & \B{30.6} \sd{.1}& 23.0 \sd{.1}    & 0.59 \sd{.01}    & \B{51.4} \sd{.1}& \B{0.60} \sd{.000} \\
    \cmidrule(l){2-6}
    & \MC{5}{l}{\tableh{\vatex\ Video$\rightarrow$Chinese ($C=10$)}}\\
    \textsc{RvR} &17.3 \sd{1.5}    & 22.1 \sd{.7}    & 0.42 \sd{.01}     & 45.8 \sd{1.4}    & 0.56 \sd{.015} \\
    \textsc{SvR} &\B{25.3} \sd{.7}& \B{24.5} \sd{.3}& \B{0.44} \sd{.01} & \B{50.8} \sd{.5}& \B{0.59} \sd{.005} \\
    \bottomrule
    \end{tabular}%
    }
    \caption{Leave-one-out averages for REF vs.\ REF (RvR) and SYS vs.\ REF (SvR) evaluations.}
    \label{tbl:gold_vs_gold}
  }
  \quad\,\,
  \parbox[t]{.41\columnwidth}{
    \centering
    \resizebox{.416\columnwidth}{!}{%
    \begin{tabular}{@{}lccccc@{}}
    \toprule
    $\rightarrow$\textsc{Unk} &
    \MC{1}{c}{\textsc{Bl}}   &
    \MC{1}{c}{\textsc{Mt}}   &
    \MC{1}{c}{\textsc{Cr}}   &
    \MC{1}{c}{\textsc{Rg}}   &
    \MC{1}{c}{\textsc{Bs}}
    \\ \midrule
    -- & 19.95 & 24.41  & 0.896 & 47.4 & 0.551 \\ \midrule
    people &19.53 & 24.24  & 0.878 & 47.0 & 0.547 \\
    (0.61\%)   &\da{.42} & \da{.17} & \da{.018} & \da{.4} & \da{.004} \\
    \midrule
    standing & 19.45 & 24.12 & 0.869 & 47.0 & 0.542 \\
    (0.88\%)     & \da{.50}  & \da{.29}  & \da{.027} & \da{.4} & \da{.009} \\
    \midrule
    sitting & 19.11 & 24.02 & 0.859 & 47.0 & 0.540 \\
    (1.06\%)  & \da{.84}  & \da{.39}  & \da{.037} & \da{.4} & \da{.011} \\
    \midrule
    man    & 19.11 & 23.66 & 0.859 & 46.6 & 0.539 \\
    (1.21\%) & \da{.84}  & \da{.75}  & \da{.037} & \da{.8} & \da{.012} \\
    \midrule
    a      &  9.20 & 20.03 & 0.490 & 34.8 & 0.480 \\
    (16.7\%) & \da{10.75} & \da{4.38}  & \da{.406}  & \da{12.6} & \da{.071} \\
    \midrule
    \midrule
    \textsc{Random} & 13.63     & 19.58      & 0.500      & 41.7     & 0.413  \\
    (16.7\%)        & \da{6.32} & \da{4.83}  & \da{.396}  & \da{5.7} & \da{.138}  \\
    \bottomrule
    \end{tabular}%
    }
    \caption{Frequent $1$-gram substitutions: \% is the percentage of tokens mapped to \T{UNK}.}
    \label{tbl:ngram_diffs}
  }
\end{table}

\paragraph{Tasks.} We explore two visual captioning tasks, namely, COCO English image captioning~\cite{mscoco} 
and the bilingual \vatex\ video captioning dataset~\cite{wang-vatex-2019}. For the former, we use a state-of-the-art captioning model~\cite{lu2018neural}, and evaluate the 5k sentences in the test split~\cite{karpathy2015deep}. For English and Chinese \vatex , we train neural baselines and evaluate on the \textit{dev} set which contains 3k videos. Finally, we also experiment with the Chinese$\rightarrow$English NIST 2008~(MT08) test set which contains 1357 sentences, and translate it using \textit{Google Translate}.

\paragraph{Results.} Table~\ref{tbl:gold_vs_gold} shows that human-authored references obtain relatively worse scores than systems in general. \textsc{Bleu} exhibits the biggest differences, with almost $11$ points for COCO and $3.5$ points for the NIST MT experiment. The results are consistent across different datasets and languages except for English \vatex\ and NIST-2008, which score slightly better than system outputs in \textsc{Meteor} and \textsc{Cider}. Overall, the results suggest that the metrics do not necessarily reflect the quality of the generated captions or translations, as
scores for human-authored texts are far from the upper bounds of the metrics (\eg 100 for \textsc{Bleu}). Therefore, we suggest that this type of evaluation should not be used to draw conclusions on `human parity', as previously done in some studies~\cite{xu2015show,vinyals2015show}.

\subsection{$N$-gram perturbations}
\label{sec:perturb}
In this section, we propose two perturbation experiments to investigate the sensitivity of metrics to frequent and infrequent phenomena. In both experiments, the first reference corpus $R_1$ of COCO test set is considered to be the output of a hypothetical system from which perturbed versions are created. Each version is then evaluated against the four remaining references $\{R_2,\dots,R_5\}$ of the COCO test set. Although we target unigrams for perturbation, they implicitly effect higher order $n$-grams as well.

\subsubsection{Frequent unigram perturbation}
\label{sec:freqpert}

We create five independent variants of $R_1$ by substituting the most common words \textit{`people', `standing', `sitting', `man'} and \textit{`a'} with the unknown token \T{UNK}. Table~\ref{tbl:ngram_diffs} shows that metrics react quite conservatively to this attack, until very high substitution percentages are reached. For example, substituting `people' yields a drop of less than $0.4$ points for \textsc{Bleu, Meteor} and \textsc{Rouge}, a potentially uninteresting drop which can easily be overlooked during model development.

Although repeatedly missing indefinite articles `a' is \I{semantically} much less critical than missing `people', all metrics aggressively penalise the former, simply because the function word `a' is extremely frequent: with 53.8\% relative drop, \textsc{Bleu} is the most affected metric, whereas \textsc{Meteor} and \textsc{BertScore} are more robust with relative drops of 17.9\% and 12.9\%, respectively. To find out more, we randomly substitute content words from $R_1$ (\textsc{Random} in Table~\ref{tbl:ngram_diffs}), until we reach the same percentage (16.7\%) as `a'. The results are interesting: 
\textsc{Bleu} (9.20$\rightarrow$13.63) and \textsc{Rouge} (34.8$\rightarrow$41.7) show substantial increases when compared to dropping `a', supporting our \I{semantic} conjecture above. However, the decreases in \textsc{Meteor} (20.03$\rightarrow$19.58) and \textsc{BertScore} (0.480$\rightarrow$0.413) highlight the sensitivity of these metrics to semantics
rather than pure frequencies. The continuous and contextual nature of \textsc{BertScore} seems to be an advantage here.


\subsubsection{Insensitivity to infrequent constructions}
\label{sec:infreq}
We now approach to the problem from the other end to explore how sensitive  metrics are to a set of rarely occurring words. We conjecture that this will provide insight into how much the metrics reward a hypothetical model that systematically translates a set of rare words correctly.
Specifically, we create four variants of $R_1$, with each one substituting a particular set of words by \T{UNK}, based on training set frequencies.\footnote{This simulates the construction of a training vocabulary by only retaining tokens that occur at least $T$ times \ie short-listing.}
Table~\ref{tbl:unkdiffs} shows that even the most aggressively short-listed hypothetical model ($T=30$)
obtains marginally worse scores than the full vocabulary model ($T=1$).
As a concrete example, the last row shows the non-substantial impact of systematically replacing each occurrence of `woman' (0.54\% of all unigrams) with `man' (1.21\% of all unigrams).

\paragraph{Discussion.}
Although it is hard to interpret the magnitude of differences observed in both experiments, our empirical findings highlight interesting cases. Overall, the conclusions of our perturbation experiments are in line
with the concurrent work of \citet{mathur-etal-2020-tangled} which suggests that important conclusions, such as comparative judgments about systems, should not be drawn based only on small changes in automatic metrics.



\begin{table}[t]
  \parbox[t]{.52\columnwidth}{
  \centering
  \resizebox{.515\columnwidth}{!}{%
  \begin{tabular}{@{}llrrrrr@{}}
  \toprule
  $\ge T$ & \T{UNK} & \MC{1}{c}{\textsc{Bl}}  &
       \MC{1}{c}{\textsc{Mt}} & 
       \MC{1}{c}{\textsc{Cr}} &
       \MC{1}{c}{\textsc{Rg}} & 
       \MC{1}{c}{\textsc{Bs}}
       \\ \midrule
  30 & 1.44\% & 19.80 & 24.25 & 0.882 & 47.2  & 0.543  \\ 
  20 & 1.07\% & 19.84 & 24.30 & 0.887 & 47.3  & 0.545  \\ 
  10 & 0.67\% & 19.89 & 24.34 & 0.891 & 47.3  & 0.548  \\  
  5 & 0.41\% & 19.91 & 24.37 & 0.893 & 47.3  & 0.549   \\ 
  1 & 0\% & 19.96 & 24.41 & 0.896 & 47.4  & 0.551  \\
  \midrule
  woman & $\rightarrow$ man & 19.63  & 24.15  & 0.889 & 47.0 & 0.549 \\
  \bottomrule
  \end{tabular}%
  }
  \caption{Insensitivity to infrequent words: T is the occurrence threshold used to obtain the vocabularies.}
  \label{tbl:unkdiffs}
  }
  \hfill
  \parbox[t]{.45\columnwidth}{
    \centering
    \resizebox{.396\columnwidth}{!}{%
    \renewcommand*{\arraystretch}{1.0}
    \begin{tabular}{@{}lrrc@{}}
    \toprule
    &\textsc{Train} & \textsc{Test} & \textsc{Refs} \\ \midrule
    \textsc{Msvd}        & 39K     & 670   & 41        \\
    \textsc{DailyDialog} & 87K     & 6.7K  & 5         \\
    \textsc{Flickr30k}   & 149K    &  1K   & 5         \\
    \vatex\              & 259K    & 3K    & 10        \\
    \textsc{Coco}        & 550K    & 5K    & 5         \\
    \textsc{Mt}          & 1.6M    & 1.3K    & 4         \\
    \bottomrule
    \end{tabular}%
    }
    \caption{Dataset statistics: unique counts are given for training set sentences.}
    \label{tbl:datastats}
  }
\end{table}

\subsection{Single representative sentence}
\label{sec:bss}
Following the observations from the previous experiments, we search over the training set for a \textit{single representative sentence} (SS) which maximises test set \textsc{Bleu}\footnote{Maximising other metrics could lead to finding other sentences and therefore different results.} when used as a system output for \textit{every} test set instance.\footnote{We note that the retrieved sentences almost never occur in test set references.}
We explore \B{tasks} and \B{datasets} which include the ones previously introduced ($\S$~\ref{sec:humaneval}). For visual captioning, we add MSVD~\cite{chen-dolan-2011-collecting}, a widely known dataset of 1,970 videos, with up to 41 English captions per video. For image captioning, we use English Flickr30k~\cite{young-etal-2014-image}, and the STAIR~\cite{yoshikawa-etal-2017-stair} dataset which provides Japanese captions for COCO images. We also explore the multi-turn dialogue dataset DailyDialog~\cite{li-etal-2017-dailydialog}
which contains conversations that cover 10 different daily life topics,
and its multi-reference test set~\cite{gupta2019investigating}.
Table~\ref{tbl:datastats} summarises the statistics about the datasets explored for this experiment.

Table~\ref{tbl:bss_scores}
draws a comparison between the scores obtained for the retrieved single sentences, baselines and state-of-the-art systems when available.
We observe that: (i) \textsc{Msvd} exhibits the highest scores with 30.6 \textsc{Bleu} and 23.4 \textsc{Meteor}, the latter being very close to a strong captioning baseline~\cite{venugopalan2015sequence}, (ii) the SS scores for the \textsc{DailyDialog} are surprisingly on par with a recent baseline and a state-of-the-art system, and (iii) \textsc{Cider} is more robust against the single sentence adversary as more references seem to affect its internal consensus-based scoring.
Although \textsc{BertScore} does not explicitly rely on $n$-gram statistics, it also exhibits surprisingly high scores for the SS setting. This contradicts the salient claims regarding the utility of the metric for model selection~\cite{bertscore}.
\begin{table*}[t!]
\centering
\resizebox{.999\textwidth}{!}{%
\begin{tabular}{@{}clcccccc@{}}
\toprule
\textsc{Dataset} &
\textsc{Method} &
\textsc{Bl1} &
\textsc{Bl4} &
\textsc{Mt} &
\textsc{Cr} &
\textsc{Rg} &
\textsc{Bs}
\\ \midrule
\textsc{Coco-En} & \BSSCocoEn                                 & 46.4      & 9.1       & 10.5      & 0.07      & 34.8 & 0.373 \\
               & Neural Baby Talk~\cite{lu2018neural}       & 75.2      & 34.4      & 26.5      & 1.06      & 55.5 & 0.628 \\
\midrule
\textsc{Flickr-En}    & \BSSFlickr                            & 46.2      & 12.2      & 11.6      & 0.07      & 32.7 & -- \\
               & Hard Attention~\cite{xu2015show}      & 66.9      & 19.9      & 18.5      & --        & --   & -- \\
                      & Neural Baby Talk~\cite{lu2018neural}  & 69.0      & 27.1      & 21.7      & 0.58      & --   & -- \\
\midrule
\textsc{Msvd-En}         & \BSSMSVD                                     & 73.8      & 30.6      & 23.4      & 0.15      & 59.7 & -- \\
        & \cite[FLOW]{venugopalan2015sequence}         & --        & --        & 24.3      & --        & --   & -- \\
                         & State-of-the-art~\cite{aafaq2019spatio}      & --        & 47.9      & 35.0      & 0.78      & 71.5 & -- \\
\midrule
\vatex-\textsc{En} & \T{[test]} \BSSVatexEn                        & 54.2      & 14.0      & 13.6      & 0.05      & 34.6 & -- \\
        & \T{[test]} Baseline~\cite{wang-vatex-2019}    & 71.3      & 28.5      & 21.6      & 0.45      & 47.0 & -- \\
                   & \T{[test]} Leaderboard Winner                 & 81.9      & 39.1      & 25.8      & 0.73      & 53.3 & -- \\
\cmidrule(l){2-8}
                   & \T{[dev]} \BSSVatexEn                         & 57.3      & 12.9      & 14.5      & 0.05      & 38.9 & 0.500 \\
                   & \T{[dev]} Baseline~\cite{wang-vatex-2019}     & 76.3      & 32.0      & 23.3      & 0.59      & 52.1 & 0.605 \\
\midrule
\textsc{DDialog-En} & \BSSDDialog                          & 36.5      & 3.8       & 10.9      & 0.04      & 24.3 & 0.225  \\
        & Seq2Seq~\cite{gupta2019investigating}   & --        & 4.5       & 9.7       & --        & 29.3 & --  \\
                 & SoTA~\cite[HRED]{gupta2019investigating}& 45.8      & 6.2       & 11.1       & 0.10     & 32.9 & 0.289  \\
\midrule
\textsc{Coco-Ja}& \BSSCocoJa                                            & 51.5      & 16.3      & 17.8      & 0.08      & 37.2 & -- \\
        & EN caption + MT~\cite{yoshikawa-etal-2012-sentence}   & 56.5      & 12.7      & --        & 0.32      & 44.9 & -- \\
                & End-to-end~\cite{yoshikawa-etal-2012-sentence}        & 76.3      & 38.5      & --        & 0.83      & 55.3 & -- \\
\midrule
\vatex-\textsc{Zh} & \T{[test]} \BSSVatexZh                        & 70.8      & 21.7      & 27.7      & 0.06      & 48.6 & --\\
        & \T{[test]} Baseline~\cite{wang-vatex-2019}    & 74.5      & 24.8      & 29.4      & 0.35      & 51.6 & --\\
                   & \T{[test]} Leaderboard Winner                 & 83.0      & 32.6      & 32.5      & 0.64      & 56.7 & --\\
\cmidrule(l){2-8}
                   & \T{[dev]} \BSSVatexZh                         & 68.2      & 19.2      & 22.5      & 0.07      & 45.8 & 0.531   \\
                   & \T{[dev]} Baseline~\cite{wang-vatex-2019}     & 75.4      & 26.8      & 25.0      & 0.44      & 51.7 & 0.597 \\
\bottomrule
\end{tabular}%
}
\caption{Comparison of single sentence scores to several baselines and state-of-the-art: \bss{Single representative sentences are in italics.} English translations are provided for non-English tasks.}
\label{tbl:bss_scores}
\end{table*}

\section{Discussion}
\label{sec:discussions}
In this work, we explore cases where commonly used language generation evaluation metrics exhibit \textit{counter-intuitive} behaviour. Although the main goal in language generation tasks is to generate `human-quality' texts, our analysis in $\S$~\ref{sec:humaneval} shows that metrics have a preference towards machine-generated texts rather than human references.
Our perturbation experiments in $\S$~\ref{sec:infreq} highlight potential insensitivity of metrics to lexical changes in infrequent $n$-grams. This is a major concern for tasks such as multimodal machine translation~\cite{specia-etal-2016-shared} or pronoun resolution in MT~\cite{guillou-etal-2016-findings}, where the metrics are expected to capture lexical changes which are due to rarely occurring linguistic ambiguities. We believe that targeted probes~\cite{isabelle-etal-2017-challenge} are much more reliable than sentence or corpus level metrics for such tasks.
Finally, we reveal that metrics tend to produce unexpectedly high scores when each test set hypothesis is set to a particular training set sentence, which can be thought of as finding the \textit{sweet spot} of the corpus ($\S$~\ref{sec:bss}). Therefore, we note that a high correlation between metrics and human judgments is not sufficient to characterise the reliability of a metric. The latter probably requires a thorough exploration and mitigation of adversarial cases such as the proposed \textit{single sentence} baseline.

\section*{Acknowledgments}
We thank the anonymous AACL-IJCNLP and COLING reviewers whose suggestions helped refine the paper.
The authors received funding from the MultiMT (H2020 ERC Starting Grant No. 678017) and MMVC (Newton Fund Institutional Links Grant, ID 352343575) projects.

\bibliographystyle{coling}
\bibliography{coling2020}

\end{document}